\documentclass[letterpaper, 10 pt, conference]{ieeeconf}  % Comment this line out if you need a4paper

\usepackage[font=small, labelfont=bf]{caption}
\usepackage{multicol}
\makeatletter
\let\NAT@parse\undefined
\makeatother
\usepackage[colorlinks = true,
            linkcolor = blue,
            urlcolor  = blue,
            citecolor = blue,
            anchorcolor = blue]{hyperref}
\usepackage{lipsum}
\usepackage{xspace}
\usepackage{amsmath}
\usepackage{amssymb}
\usepackage{booktabs} 
\usepackage{wrapfig}
\usepackage[capitalise]{cleveref}
\usepackage[english]{babel}
\usepackage{here}
\usepackage{flushend}
\usepackage[colorinlistoftodos]{todonotes}
\usepackage{multirow}
\usepackage{dsfont}
\usepackage{subcaption}
\usepackage{graphicx} 
\usepackage{mathtools}
\usepackage{adjustbox}
\usepackage{booktabs, makecell, tabularx}
\usepackage{cite}

\IEEEoverridecommandlockouts                              % This command is only needed if 
\title{\LARGE \bf
Learning from Demonstration with Hierarchical Policy Abstractions Toward High-Performance and Courteous Autonomous Racing
}
\author{Chanyoung Chung$^{1\dagger}$, Hyunki Seong$^{2\dagger*}$, David Hyunchul Shim$^{2}$ \\
\thanks{
This work was supported by the Technology Innovation Program (Development of drone-robot cooperative multimodal delivery technology for cargo with a maximum weight of 40kg in urban areas) funded by the Ministry of Trade, Industry \& Energy (MOTIE), South Korea, under Grant RS-2023-00256794.
}
\thanks{$^\dagger$Equally contributed. $^*$Corresponding author.}
\thanks{$^{1}$ Chanyoung Chung is with the Field AI, Mission Viejo, CA 92691 USA.
        {\tt\small calvin.chanyoung@gmail.com}
        }%
\thanks{$^{2}$Hyunki Seong and David Hyunchul Shim are with the School of Electrical Engineering, Korea Advanced Institute of Science and Technology, Daejeon, South Korea.
        {\tt\small \{hynkis,hcshim\}@kaist.ac.kr}
        }%
}
\begin{document}

\maketitle
\thispagestyle{empty}
\pagestyle{empty}

%%%%%%%%%%%%%%%%%%%%%%%%%%%%%%%%%%%%%%%%%%%%%%%%%%%%%%%%%%%%%%%%%%%%%%%%%%%%%%%%
\begin{abstract}
Fully autonomous racing demands not only high-speed driving but also fair and courteous maneuvers. In this paper, we propose an autonomous racing framework that learns complex racing behaviors from expert demonstrations using hierarchical policy abstractions. At the trajectory level, our policy model predicts a dense distribution map indicating the likelihood of trajectories learned from offline demonstrations. The maximum likelihood trajectory is then passed to the control-level policy, which generates control inputs in a residual fashion, considering vehicle dynamics at the limits of performance. We evaluate our framework in a high-fidelity racing simulator and compare it against competing baselines in challenging multi-agent adversarial scenarios. Quantitative and qualitative results show that our trajectory planning policy significantly outperforms the baselines, and the residual control policy improves lap time and tracking accuracy. Moreover, challenging closed-loop experiments with ten opponents show that our framework can overtake other vehicles by understanding nuanced interactions, effectively balancing performance and courtesy like professional drivers.
% Moreover, closed-loop experiments show that our framework can overtake other vehicles by understanding nuanced interactions, effectively balancing performance and courtesy like professional drivers.

% Towards the fully autonomous race, which is analogous to races between professional drivers,
\end{abstract}

%%%%%%%%%%%%%%%%%%%%%%%%%%%%%%%%%%%%%%%%%%%%%%%%%%%%%%%%%%%%%%%%%%%%%%%%%%%%%%%%
\section{Introduction}
\label{sec:intro}

% Recently, autonomous racing has received a lot of attention as a way to push the boundaries of autonomous vehicle technology. It serves as a valuable testbed to showcase the capabilities and reliability of autonomous systems under extreme conditions, including high-speed navigation and control, low-latency computation, and real-time operation.

% Real-world autonomous racing competitions, such as Roborace \cite{roborace} and the Indy Autonomous Challenge (IAC) \cite{iac}, have been at the forefront of testing high-speed autonomy. The IAC, in particular, is the world's first 1:1 overtaking competition, held at Las Vegas Motor Speedway (LVMS), where the winner was determined by the vehicle's ability to overtake the 'defending' vehicle at a higher speed. The participating teams, including the authors as part of team KAIST, successfully demonstrated high-speed passing at over 200 km/h \cite{jung2023autonomous, betz2022tum, lee2022resilient}.

Autonomous racing has recently gained attention as a way to push the boundaries of autonomous vehicle technology, serving as a testbed to showcase system capabilities under extreme conditions, such as high-speed navigation, low-latency computation, and real-time operation. Competitions like Roborace \cite{roborace} and the Indy Autonomous Challenge (IAC) \cite{iac} have led the way in testing high-speed autonomy. The IAC, the world’s first 1:1 overtaking competition held at Las Vegas Motor Speedway (LVMS), determined the winner based on the vehicle's ability to overtake a 'defending' vehicle at higher speeds. Teams, including the authors as part of team KAIST, successfully demonstrated high-speed passing at over 200 km/h \cite{jung2023autonomous, betz2022tum, lee2022resilient}.

\begin{figure}[!t]
\centering
\includegraphics[width=1.\columnwidth]{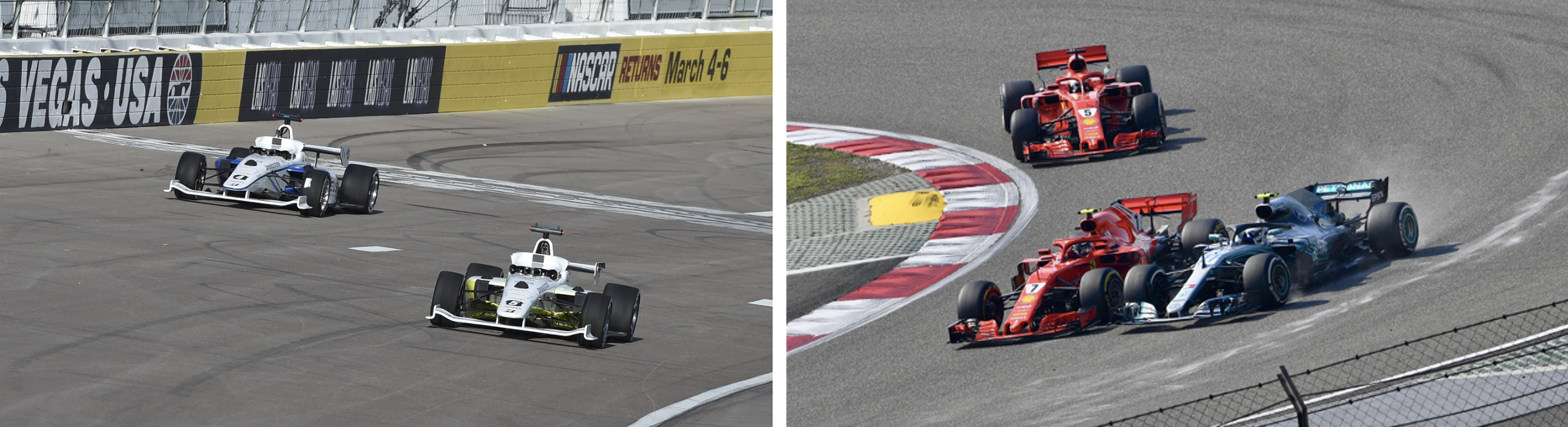}
\caption{Overtaking scenarios at the Indy Autonomous Challenge (left) and Formula 1 motorsport (right)}
\label{fig:concept}
\vspace{-2em}
\end{figure}

Despite successful demonstrations of high-speed autonomous driving, challenges remain in achieving professional human-level racing standards. Racing scenarios typically require rules for fairness and safety. For example, the IAC imposes specific rules on 'defender and offender roles,' \cite{jung2023autonomous} including target driving speeds, limited overtaking zones, and designated paths for defenders. Similarly, the Indy 500 and Formula 1 (Fig. \ref{fig:concept}) enforce rules to maintain fairness and safety, albeit with fewer restrictions. In all cases, drivers—human or autonomous—must proactively interact with others, balancing agility and courtesy in competitive, adversarial settings.
% Despite successfully demonstrating high-speed autonomous driving, challenges still remain in meeting professional human-level racing standards. Typical racing scenarios require rules to ensure fairness and driver safety during high-speed driving.  Standard racing scenarios usually require rules to ensure fairness and driver safety during high-speed driving. For example, the IAC imposes specific rules on 'defender and offender roles,' including target driving speeds, limited overtaking zones, and designated paths for defenders. Similarly, the Indy 500 and Formula 1 implement rules to maintain fairness and safety, though with fewer restrictions. (Fig. \ref{fig:concept}). In all cases, human or autonomous drivers must proactively interact with others, balancing agility and courtesy in adversarial scenarios to ensure safe and competitive racing.
% Despite the successful demonstration of high-speed autonomous driving, there are still significant gaps when it comes to 'professional human level of racing' scenarios. For example, the IAC had specific rules regarding "defender and offender roles," which included target driving speeds, limited overtaking zones, and designated paths for the defender. In contrast, motorsports such as Formula 1 and Indy500 have minimal race rules to ensure fairness and drivers' safety (Fig. \ref{fig:concept}). Therefore, drivers should proactively interact with others, balancing agility and courtesy in adversarial scenarios to ensure safe racing.

In this study, we propose an offline learning-based autonomous racing framework using hierarchical policy abstractions. The proposed framework consists of two levels of policy abstraction: 1) a novel trajectory planning policy (TPP) and 2) a residual control policy (RCP), both trained using a learning from demonstration method (Fig. \ref{fig:overview}). The TPP predicts an optimal future trajectory using a density distribution map that considers environmental and interaction contexts with surrounding opponents in adversarial driving scenarios. Subsequently, based on the inferred trajectory, the RCP refines the control inputs from the forward controller by adding residual control adjustments. This modular policy architecture leverages the learning from demonstration paradigm to abstract expert-driving policies at hierarchical levels, effectively learning complex policies that balance performance and courtesy.

We extensively validate our proposed method using a high-fidelity racing simulator in multi-agent, highly competitive scenarios. Our open-loop results show that our trajectory-level policy outperforms baselines both quantitatively and qualitatively. Furthermore, in closed-loop experiments, our approach demonstrates strategic overtaking while considering safety and courtesy in response to nuanced interactions with surrounding opponents. These experiments highlight the effectiveness of our autonomous driving framework in balancing agility and fairness during high-speed racing.
% Unlike previous works, we do not assume an advantage for the ego vehicle, such as slow or conservative opponents that simplify overtaking scenarios.
% The proposed method was extensively validated using a high-fidelity racing simulator under N-players and competitive scenarios without assuming any predominance of the ego vehicle.

% [Contribution]
The main contributions of our work are as follows:
\begin{itemize}
    \item
    We propose an autonomous racing framework using hierarchical policy abstractions that are trained through the learning from demonstration paradigm.
    % , which aligns with the classical autonomy pipeline.
    % Autonomous racing framework using hierarchical policy abstraction with the "learning from demonstration" paradigm, which aligns with the classical autonomy pipeline is proposed.
    % Our design choice leverages the "learning from demonstration" paradigm without compromising the benefits of the classical approach.
    % \item Offline RL-based autonomous racing framework using hierarchical policy abstraction, which aligns with the classical autonomy pipeline is proposed. Our design choice leverages the "learning from demonstration" paradigm without compromising the benefits of the classical approach.
    % \item Density estimator-based neural network is proposed for the trajectory-level policy, which can reason about its output based on the likelihood of prior distribution. Proposed policy models are carefully designed to run at a high inference rate (over 50Hz) considering the application domain.
    \item We design a density estimator-based trajectory planning policy and a residual control policy, both of which reason about their outputs at different levels of abstraction.
    % The proposed policy models are carefully designed to run at a high inference rate (over 50Hz) considering the application domain.
    \item We extensively evaluate our approach in a high-fidelity racing simulator with challenging multi-agent adversarial scenarios, demonstrating well-balanced performance in agility, safety, and courtesy during high-speed racing.
\end{itemize}

% [Paper structures]

\section{Related Work}
\label{sec:related_work}

\subsection{Model-Predictive Control}

% The MPC, which aims to find the sequence of optimal control commands with non-linear constraints and dynamics, is one of the most well-studied frameworks for autonomous racing applications \cite{kalaria2021local, you2019high, zhang2018tire, wischnewski2022tube}. In \cite{kabzan2020amz}, they formulated the objective function of MPC which aims to maximize the progress along the reference path by integrating a non-linear vehicle model for time trial racing. In \cite{liniger2015optimization}, they proposed a model predictive contouring control (MPCC) approach that can follow the reference path and avoid stationary obstacles. They demonstrated the real-time capability of the proposed method through real-world experiments using a 1:43 scale vehicle. In \cite{dixit2019trajectory}, overtaking decisions and planning are integrated into the objective function of MPC. By transforming the motion planning problem into a finite-time quadratic programming problem and modeling the non-convex collision avoidance constraints, they demonstrated high-speed overtaking with one traffic participant. MPC showed promising results in various high-speed autonomous driving scenarios; however, the literature mainly focused on pushing the vehicle to the limit of the dynamics. Few studies tackled the multi-agent scenarios by incoporating collision and overtaking behavior into the objective function, but their scenarios is simple compares to the realistic racing scenarios, for example, only one traffic participant or slow moving vehicle and so on. 

Model Predictive Control (MPC) is a popular planning and control framework in high-speed driving applications, which determines the optimal sequence of control input while taking into account nonlinear and dynamic constraints \cite{kalaria2021local, you2019high, zhang2018tire, wischnewski2022tube}. For the time trial racing without any other agents, the objective function of MPC can be formulated to maximize progress along a reference path using a nonlinear vehicle model \cite{liniger2015optimization,kabzan2020amz}.
Several studies in urban driving scenarios incorporate risk evaluation as either a constraint \cite{bernhard2021risk} or an objective function \cite{zhang2024courteous} to address safety and courtesy. While MPC has shown promising results in autonomous driving, many previous studies rely on simplified scenarios and may potentially face significant computational and stability issues due to constrained optimization.
% In \cite{liniger2015optimization}, a model predictive contour control was adapted that could track a predefined reference path and avoid stationary obstacles; this approach was tested in real-world experiments using a small-scale vehicle.
% By transforming the motion planning problem into a finite-time quadratic programming problem and modeling non-convex collision avoidance constraints, \cite{dixit2019trajectory} incorporated overtaking decisions and planning into the MPC objective function, achieving high-speed overtaking with one traffic participant.
% While MPC has shown promising results in high-speed autonomous driving where understanding dynamics becomes critical, a large body of previous studies relies on simplified driving scenarios and potentially contain large computation and stability issues because of constrained optimization.

\subsection{Game-Theoretic Planning}
% wang2019game,liniger2019noncooperative,jung2021game
% Game theory is used to model decision-making in strategic situations where different individuals or groups have conflicting interests. It provides a framework for analyzing and understanding the interdependent decisions of rational individuals. The authors of \cite{wang2019game} proposed a game-theoretic planner for overtaking in a two-car racing scenario. They used an iterative best response algorithm (IBR) that seeks for the Nash equilibrium within the joint trajectory of the two vehicles. Similarly, in \cite{liniger2019noncooperative}, a game-theoretic approach was proposed to model racing as a non-cooperative non-zero-sum game, while assuming open-loop information structures. In our previous study \cite{jung2021game}, we introduced a game-theoretic MPC for head-to-head autonomous racing, and we modeled the competitive overtaking situation with n-players into a series of Stackelberg game, a 2-player game. The performance of the proposed method was demonstrated during the IAC simulation race under competitive racing scenarios. However, solving the game in real-time fashion is not computationally feasible especially in the multi-agents scenarios. 

Game theory models decision-making in situations with opposing interests, making it a valuable tool for developing adversarial driving strategies. In \cite{wang2019game}, a game theory-based trajectory planning algorithm was proposed for two-vehicle passing scenarios, using an iterative best response algorithm to find the Nash equilibrium between trajectories. Similarly, \cite{liniger2019noncooperative} addressed the overtaking problem as a non-cooperative, non-zero-sum game with open information structures. In our previous research \cite{jung2021game}, we developed a game-theoretic control strategy for head-to-head autonomous racing by transforming a multi-player overtaking scenario into a sequence of two-player Stackelberg games. These game-theoretic interactions facilitate strategic, safe, and courteous driving in urban scenarios \cite{speidel2019towards, wang2019enabling}. However, the computational complexity of these algorithms increases exponentially with the number of players due to the nature of these algorithms.
% Game theory models decision-making in situations where different individuals or groups hold opposing interests and is a valuable tool for formulating adversarial driving. In \cite{wang2019game}, the authors proposed a game theory-based trajectory planning algorithm for two-vehicle passing scenarios, using an iterative best response algorithm to identify the Nash equilibrium between the trajectories. Similarly, in \cite{liniger2019noncooperative}, the authors modeled the overtaking problem as a non-cooperative, non-zero-sum game with open information structures. In our previous research \cite{jung2021game}, we proposed a game-theoretic control strategy for head-to-head autonomous racing that transformed a multi-player overtaking scenario into a sequence of 2-player Stackelberg games. However, computation complexity grows exponentially with the number of players due to the nature of algorithms. To address this problem, the literature has attempted to divide the multi-player game into sub-games or assume the accessibility of others' utility or payoffs, but these approaches are not always feasible in extremely competitive racing scenarios.

\subsection{Learning-based Approaches}
% One of the most relevant studies on the racing application was published in \cite{song2021autonomous}. The authors proposed a three-stage curriculum RL framework for autonomous overtaking in head-to-head racing. They extended the work in \cite{fuchs2021super}, designed for high-speed driving, to the more complex and challenging overtaking domain by redesigning the reward function to include the collision penalty. \cite{wurman2022outracing} is one of the most recent research conducted by the Sony AI group. Sophy is a racing AI agent that has learned to master the PlayStation game and driving simulator, Gran Turismo(GT) Sport. It learns through deep reinforcement learning techniques to race at the highest level. It utilizes multiple factors(progress, off-course penalty, wall penalty, tire-slip penalty, passing bonus, collision penalty, rear-end penalty, and unsporting-collision penalty) in the reward function. Sophy outperformed all the professional game players in every time-trial race. Also, it won a head-to-head competition against four of the world's best Gran Turismo drivers. Learning based approaches showed promising performance. Majority of the previous studies adapted the \emph{end-to-end} approach, which directly maps observations to control outputs; however, this \emph{end-to-end} design can be impractical when it comes to the field robotics. 

Recently, end-to-end reinforcement learning (RL) has proven effective in enabling interactive and agile vehicle control across various scenarios, including urban driving \cite{seong2021learning, yan2021courteous}, overtaking \cite{song2021autonomous}, and high-speed racing \cite{fuchs2021super}. For interactive intersection scenarios, attention mechanisms have been used to assess the relative importance of surrounding vehicles \cite{seong2021learning}, and priority-based discrete action representations have been proposed to facilitate courteous driving at unsignalized intersections \cite{yan2021courteous}.
In \cite{song2021autonomous}, a three-stage curriculum RL framework was introduced for autonomous overtaking, extending previous high-speed driving work \cite{fuchs2021super}. This framework improves overtaking by incorporating a collision penalty into the reward function. Another example is Sophy, a racing AI agent trained with deep reinforcement learning for Gran Turismo Sport \cite{wurman2022outracing}. Sophy's reward function includes factors like progress, off-course penalties, wall collisions, tire slip, unsporting behavior, and a passing bonus. 
% Recently, reinforcement learning (RL) has proven to be effective in enabling agile vehicle control, including overtaking and high-speed driving. In \cite{song2021autonomous}, the authors presented a three-stage curriculum RL framework for autonomous overtaking, extending the work done in \cite{fuchs2021super} for high-speed driving. This framework was designed to be more effective for the challenging task of overtaking by incorporating a collision penalty in the reward function. Another example is Sophy, a racing AI agent trained using deep reinforcement learning techniques to race in the game Gran Turismo Sport \cite{wurman2022outracing}. Sophy's reward function was designed to take into account multiple factors, including progress, penalties for off-course, wall, tire slip, and unsporting collisions, as well as a passing bonus. Sophy outperformed all professional human players in time-trial races and even won a head-to-head competition against four of the world's best e-sport drivers.
However, their end-to-end RL methods have limitations from a resilient system design perspective, as it is challenging to examine the reasoning and inference processes within these pipelines. Additionally, they require tailored reward designs to balance performance, safety, and courtesy.
In our work, we tackle these well-known issues by using hierarchical policy abstractions that follow the modular autonomy pipeline while leveraging the learning from demonstration fashion.

% While the aforementioned approaches have shown outstanding results, we argue that a direct mapping between input and output (control signals) has limitations from a resilient system design perspective, as it is challenging to reason about predictions and integrate with previous autonomy stacks.

% However, While the aforementioned approaches have shown outstanding results, their end-to-end control fashions with direct mapping between input and output have limitations from a resilient system design perspective as it is challenging to investigate reasoning and inference processes within their end-to-end pipelines.

\section{Problem Statement}
\label{sec:problem_statement}

% \begin{figure}[!t]
%     \centering
%     \includegraphics[width=.9\columnwidth]{figures/overall_v4.png}
%     \caption{Overview of our framework based on hierarchical policy abstraction level. Our framework follows the conventional modular autonomy pipeline while leveraging the demonstration data via an offline reinforcement learning paradigm. The red dashed and black lines indicate the training and inference flow, respectively.}
%     \label{fig:overview}
% \end{figure}

\begin{figure}[!t]
    \centering
    \includegraphics[width=.85\columnwidth]{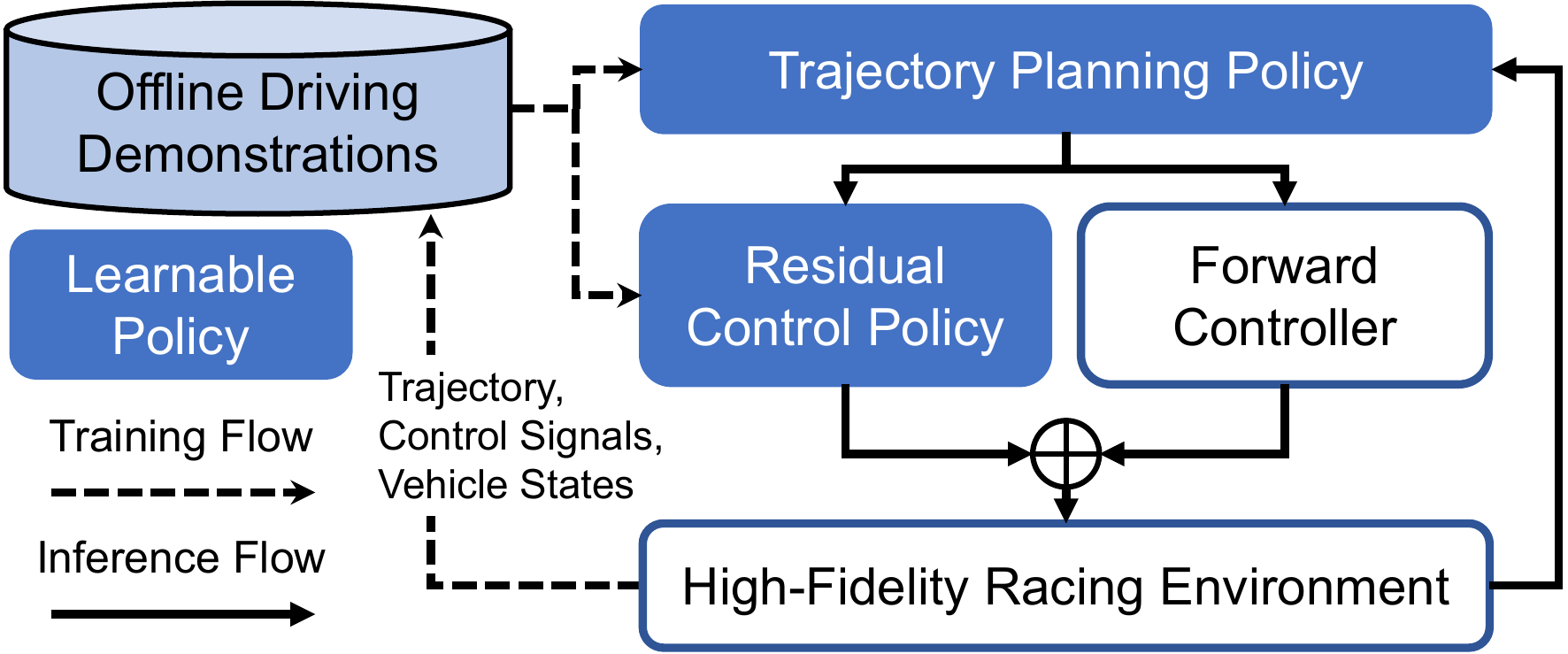}
    % \includegraphics[width=.9\columnwidth]{figures/overall_v5.png}
    % \includegraphics[width=.9\columnwidth]{figures/action_overall_v2.png}
    % Overview of action-state space level policy module.
    \caption{Overview of our autonomous racing framework.} %Our framework follows the conventional modular autonomy pipeline while leveraging the demonstration data via an offline reinforcement learning paradigm. The dashed and solid lines indicate the training and inference flow, respectively.
    \label{fig:overview}
    \vspace{-1.5em}
\end{figure}

We formulate trajectory planning as a sequential decision-making problem, following the partially observable finite Markov decision process (POMDP). Our agent's state at time t is $s_t \in \mathbb{R}^2$, where t = 0 refers to the current time step, and $\phi_t$ represents the agent's observations at time step t.

Our objective for the TPP is to predict the probability distribution of future trajectories $p(\tau_i)$, learned from the demonstration trajectories $D = \{\tau_{d,0}, \tau_{d,1}, ..., \tau_{d,m}\}$. The trajectory $\tau$ is defined as a sequence of states $((s_0, \phi_0), (s_1,\phi_1), ..., (s_N,\phi_N))$, where $s = (x, y)$, and N represents the 2D location in an ego-centric frame and the planning time horizon, respectively. The trajectory-level planning policy $\pi_p$ can be written as a function that maps from states and features to future trajectories: $\pi_p: \textbf{s}, \phi(\textbf{s})\mapsto\tau$.
After the trajectory, $\tau$ is planned, the control module is responsible for generating control inputs $U=\{u_{steer},u_{acc}\}\in\mathbb{R}^2$ to follow the trajectory precisely. The system dynamics are expressed by the probabilistic transition model, following our POMDP setup. Our residual control policy $\pi_{rc}$ aims to fill the gap between the demonstrated control, $U_d$, and the classical forward controller output, $U_f$: $\pi_{rc}: \tau\mapsto U_d-U_f$.

\begin{figure*}[t!]
    \centering
    \includegraphics[width=0.82\textwidth]{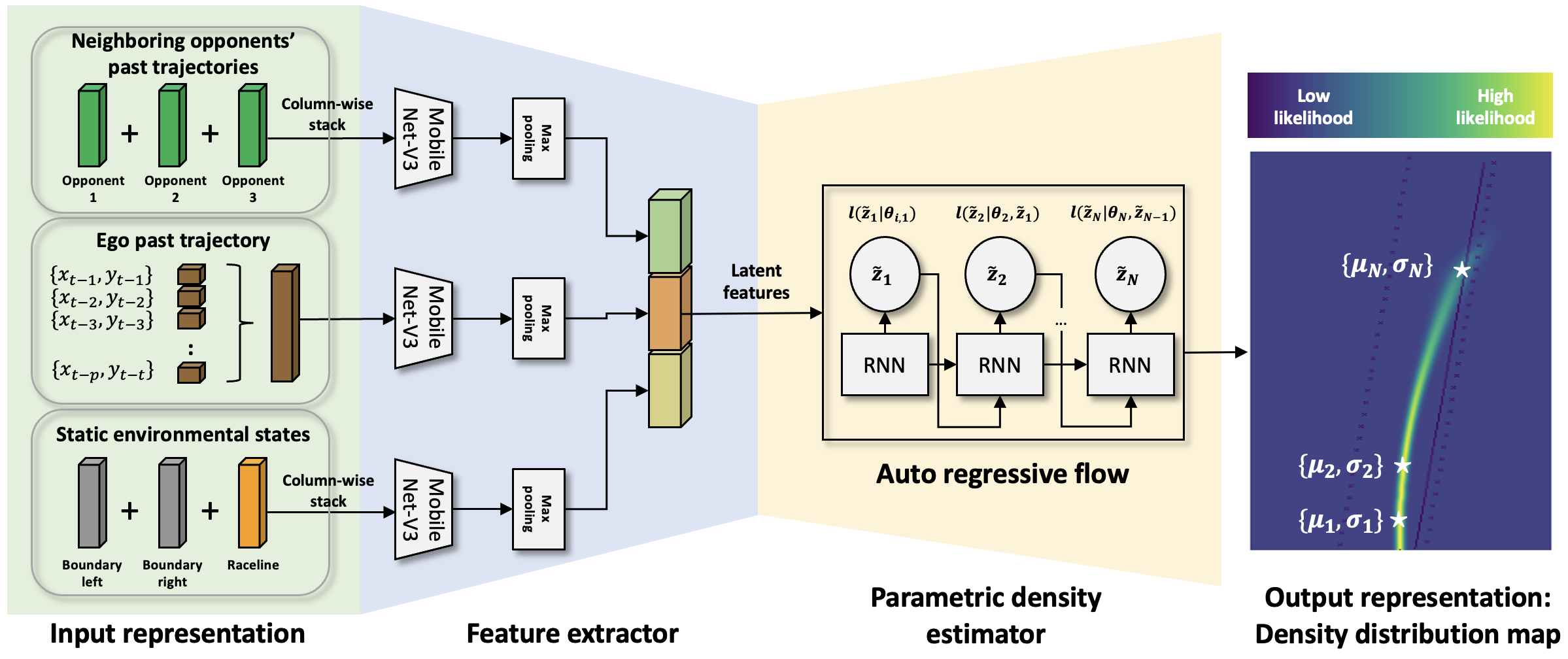}
    \caption{Illustration of our trajectory-level policy model structure. Feature extractor takes inputs about the environment, neighboring opponents, and ego past trajectory information. Extracted contextual cues are fed into the parametric density estimator using NAF, and it finally outputs the distribution of future trajectory.}
    \label{fig:net_traj}
    \vspace{-1.5em}
\end{figure*}

\section{Methodology}
\label{sec:method}
Fig. \ref{fig:overview} illustrates the overall architecture of the proposed autonomous racing framework. Unlike other learning-based autonomous racing studies, our hierarchical policy abstractions align with the modular autonomy stack, which consists of high-level planning and low-level control \cite{fleury1994design}. These hierarchical abstractions enable leveraging learning-based methods without compromising the advantages of the classical yet practical approach. In the following sections, we provide details of each level of the policy model and explain the training criteria.

\subsection{Trajectory Planning Policy}
% The goal of racing is to reach the finish line earlier than others; therefore, overtaking is one of the key tactical behaviors of racing.
Planning overtaking trajectories in adversarial scenarios requires not only avoiding collisions but also considering interactions with surrounding vehicles in terms of safety and courtesy, while ensuring the feasibility of the vehicle dynamics. To address this, we adopt an offline learning paradigm that learns from human (or expert) demonstrations, which capture complex interactions inherent in racing contexts.

% The aim of the trajectory planning policy is to reason about the planning of an expert's maneuvers, as captured by the conditional density $q(S_{1:T}|\phi) = \prod_{t=1}^T q(s_t|S_{1:t-1},\phi)$, given a fixed set of demonstrations, $\textit{D}={(s_n, \phi_n)}{n=1}^{D}$ and demonstrations are sampled from an unknown distribution of expert behavior $s_n\sim p(s{:n}|\phi_n)$. Thus, the trained policy is expected to output the density of expert-like(maximum likelihood) trajectories. 

In accordance with the problem formulation presented in Section \ref{sec:problem_statement}, we express our trajectory planning policy (TPP) model as $q(S_{1:T}|\phi)=\prod_{t=1}^T q(S_t|S_{1:t-1},\phi)$, where $\phi$ represents the learned parameters that are trained based on the distribution of expert's demonstrations $s^i\sim p(S|\phi^i)$.
During inference, TPP selects the highest-likelihood expert-like trajectory to guide the low-level control policy.
% During inference, the trajectory planning module selects the highest-likelihood expert-like trajectory, denoted as $s^* = \operatorname*{argmax}_S \log q(S|\phi)$.
% During the inference, standard objective of imitation model is to choose one high likelihood export-like trajectory.
% \begin{equation}
% s^* = \operatorname*{arg\,max}_S \log q(S|\phi).
% \end{equation}

The network architecture of TPP, shown in Fig. \ref{fig:net_traj}, includes two main components: a multiple context feature extractor and a parametric density estimator. The input is first processed by the extractor, which encodes it into a condensed latent representation. These features are then used by the conditional density estimator block to estimate the likelihood of the trajectory distribution.

The encoder comprises three branches, each dedicated to extracting contextual cues from the ego vehicle's past trajectory, environmental information, and the positions of opponents. This setup provides insights into vehicle dynamics, the static environmental context, and opponent behavior to predict their near-future actions. To enhance realism, the perception range is limited to 60 meters, and only the three nearest opponents are considered as inputs. Inputs are represented as two-dimensional XY vectors in the egocentric frame, with past trajectories segmented into 0.5-second intervals. To reduce computational complexity, the race line and track boundaries are truncated to 100 meters at 1-meter intervals. Each extractor branch uses MobileNetV3 \cite{howard2019searching} as its backbone, and the extracted latent features are concatenated into a single feature vector, which is then passed to the density estimator block.

% Density estimator 설명 + autoregressive (https://www.pnas.org/doi/10.1073/pnas.2101344118)
% https://arxiv.org/pdf/2112.09943.pdf
% To obtain an analytical expression for the probability density function (PDF) of transitions for our problem setting in Sec. \ref{sec:problem_statement}, we implemented an autoregression-based neural density estimator \cite{filos2020can, kingma2016improved}. It allows defining a parametric flow of transformations that reshapes a known initial pdf to one that best fits the expert demonstration distribution, $p_{D}$, by decomposing the density into the product of conditional densities based on the probability chain rule, $\mathds{P}(X)=\prod_i\mathds{P}(x_i|X_{1:i-1})$. Here, we modeled a conditional probability as a normal distribution model.

% We implemented a neural autoregressive flow (NAF) \cite{huang2018neural} as a parametric density estimator. NAF allows defining a parametric flow of transformations that reshapes a known initial probability density function to one that best fits the expert demonstration distribution, $p_{D}$, by decomposing the density into the product of conditional densities based on the probability chain rule, $\mathds{P}(X)=\prod_i\mathds{P}(x_i|X_{1:i-1})$. Here, we modeled a conditional probability as a normal distribution. We recommend to reader to \cite{huang2018neural,agarwal2020imitative} for the detailed theoretical background and proof. 
We employed a Neural Autoregressive Flow (NAF) \cite{huang2018neural} as a parametric density estimator. With NAF, we can define a parametric flow of transformations that modifies a known initial probability density function to better align with the expert demonstration distribution, $p_{D}$. This is achieved by breaking down the density into the product of conditional densities, which is based on the probability chain rule, $\mathds{P}(X)=\prod_i\mathds{P}(x_i|X_{1:i-1})$. Here, we modeled the conditional probability as a normal distribution. For more details on NAF, we recommend referring to \cite{huang2018neural,agarwal2020imitative}.

\subsection{Residual Control Policy} 
The selected trajectory from the planning policy is fed into the control module. Typical end-to-end supervised learning and RL-based controllers that directly output control inputs require a large number of training samples. In order to improve data efficiency, we designed our action-level policy as a residual control policy (RCP) that is trained in a supervised manner. We use the difference between the forward controller and demonstrations as the supervision signal, effectively reducing the search space of the control policy and leading to more efficient training. Specifically, we utilized the Linear Quadratic Regulator (LQR) based vehicle controller from \cite{jung2023autonomous} as the forward controller.

\begin{figure}
\centering
\centering
   \includegraphics[width=0.85\linewidth]{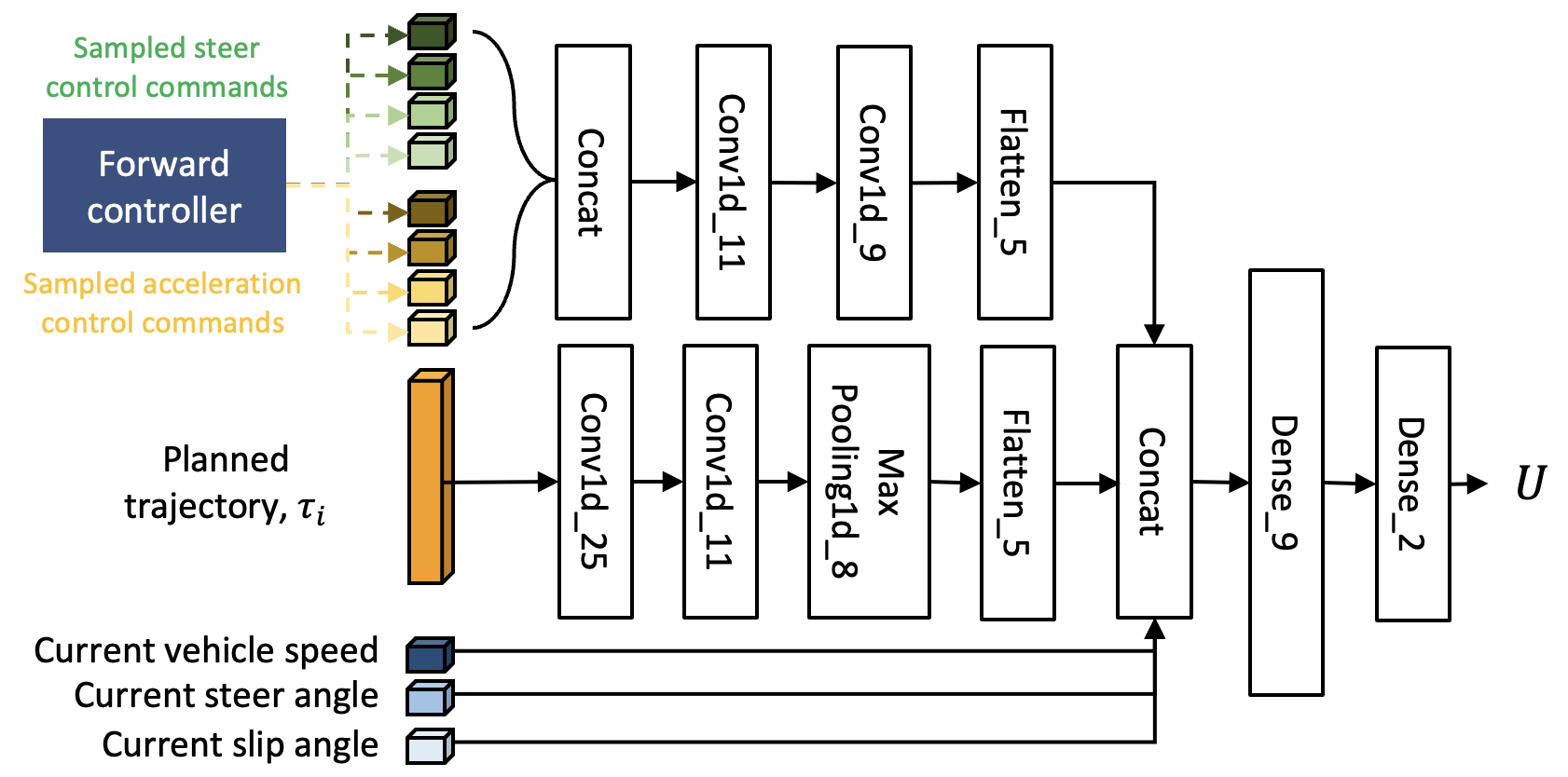}
   \caption{}
\caption[]{Action-state space level policy network architecture.}
\label{fig:action-fig}
\vspace{-1.5em}
\end{figure}

Fig. \ref{fig:action-fig} presents an overview of our overall action-level policy learning and architecture. 
The control policy model consists of fully connected layers that generate steering and acceleration control inputs, denoted by $U={u_{steer}, u_{acc}}$. The model inputs are the same as those used by the forward controller, including cross-track error, longitudinal vehicle speed, and current steering position. Additionally, sampled forward controller outputs at various look-ahead distances (4, 10, 15, and 24 meters) are provided to better understand the trajectory geometry from near to far.
% The control policy model comprises of fully-connected layers that generate steering and acceleration control inputs denoted by $U=\{u_{steer},u_{acc}\}$. The inputs of the model are the same as those used by the forward controller, including the cross-track error, longitudinal vehicle speed, and current steering position. Additionally, sampled forward controller outputs with various look-ahead distances(here, we used 4, 10, 15, and 24 meters) are provided for better understanding of the trajectory geometry from near to far.

\subsection{Loss Function Design} 
Two different levels of policies are trained separately. To train the RCP model, $\theta_a$, the L2-norm loss function is configured as follows:
\begin{align}
\begin{split}
    \mathcal{L}(a) &= 
    \lambda \vert \vert 
    (\pi_{a, steer} + \mathcal{F}_{steer}(s) - \pi_{d,steer}) 
    \vert \vert^{2}
    \\
    & +(1-\lambda) \vert \vert 
    (\pi_{a, acc} + \mathcal{F}_{acc}(s) - \pi_{d,acc})
    \vert \vert^{2},
    \label{eq:loss_action}
\end{split}
\end{align}
where $\pi_{\theta_{a,-}}$, $\mathcal{F_{-}}$ and $\pi_{d,-}$ represent action policy output, forward controller term, and control signals from the demonstration, respectively. Losses from each control modality are balanced via hyperparameter, $\lambda$. 
% loss
Weighted Maximum Likelihood Estimation (MLE) loss is employed to train the TPP model as follows:
\begin{align}
\begin{split}
    \mathcal{L}(t) 
    = - \frac{1}{N} \sum_{n=1}^{N} n \log(p(\tilde{z}_n|\theta)),
    \label{eq:general_mle}
\end{split}
\end{align}
where $N$, $\tilde{z}_n$, and $\theta$ represent the planning horizon step, policy output, and model parameters respectively. The step-wise weighted term in the loss function was introduced to prevent the model from extrapolating past trajectories without taking into account the relevant contextual information.

% For trajectory planning policy learning, we adapted simplified reverse KL as follows:
% \begin{align}
% \begin{split}
%     \mathcal{L}(b) 
%     = - \log(J(z_n|b))-\sum_{n=1}^N\log\det\abs*{\frac{dz_n}{dz_{n-1}}},
%     \label{eq:general_mle}
% \end{split}
% \end{align}
% where 

% Since our target application is high-speed driving scenarios, the beginning part (close points from the current position) of the sample usually can be cheated by just extrapolating the past trajectory without considering the contextual understanding. To precisely predict the overall planning horizon, we designed the time-weighted likelihood loss function for training as in \ref{eq:weighted_mle} and the Fig. \ref{} clearly shows its effect.

\section{Implementation and Experiment}
\label{sec:experiments}

\subsection{Simulation Environment}
\begin{figure}[t]
\centerline{\includegraphics[width=0.70\columnwidth]{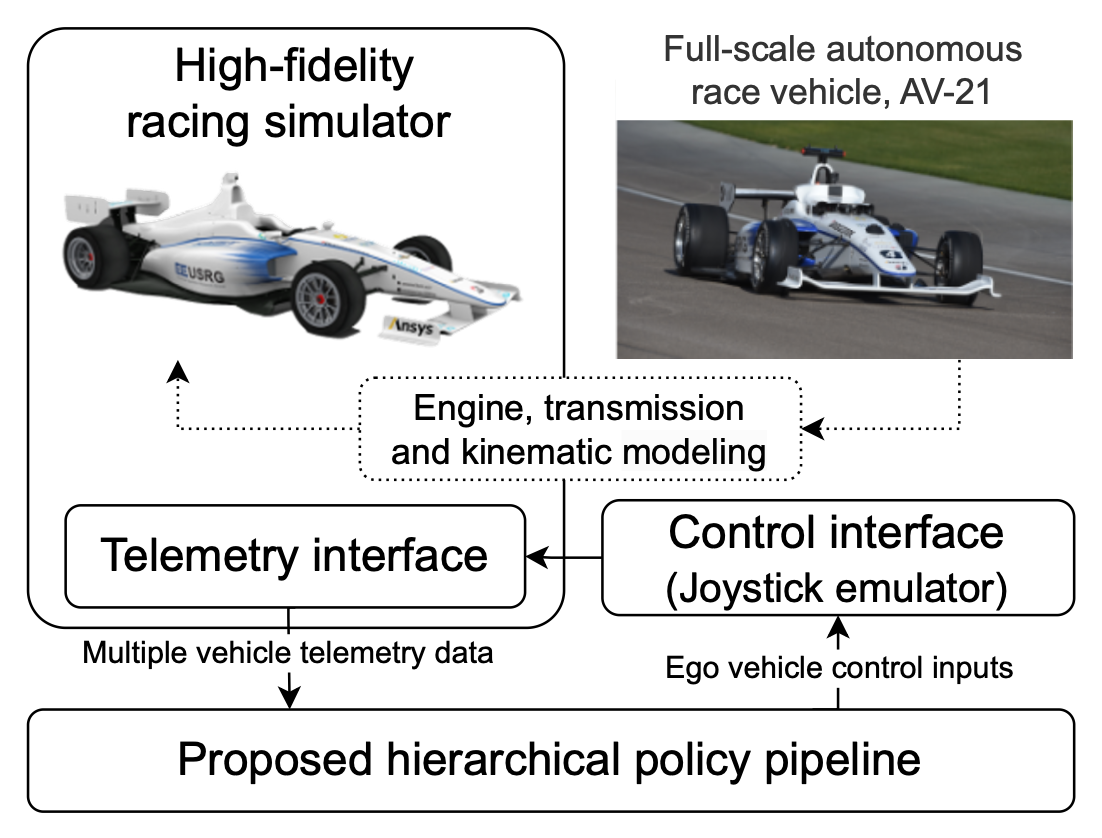}}
% \caption[Simulation environment for collecting the demonstrations and validation. Based on the real-world race vehicle spec,  powertrain modeling (engine and transmission) and chassis modeling were performed, and information necessary for head-to-head racing demonstration and autonomy stack was provided using telemetry information provided by Assetto Corsa. ]
\caption[]{Simulation environment for data collection and validation.} % Simulation vehicle is modeled after the AV-21, real-world autonomous race vehicle for IAC.
\label{fig:ac_system_diagram}
\vspace{-1.5em}
\end{figure}

To collect the training dataset and evaluate the proposed approach, we developed a racing simulator using Assetto Corsa, a well-known racing game renowned for its realistic vehicle dynamics and customization capabilities. Fig. \ref{fig:ac_system_diagram} shows our simulation environment architecture. We implemented a bi-directional interface based on UDP communication and a joystick emulator to receive live telemetry and transmit user commands. The simulation vehicle models, including the engine, transmission, and kinematics, were based on the AV-21, a full-scale autonomous race vehicle designed for the IAC.
To create adversarial scenarios, we assumed that all vehicles on the track have the same engine power, wheel torque, and maximum speed. As a result, overtaking other vehicles was only possible if opponents made mistakes or if drivers took advantage of the slipstream by closely following or staying ahead of other vehicles. This mirrors professional human racing, enhancing the simulation’s realism.
% To collect the training dataset and evaluate the proposed approach, we developed a racing simulator using Assetto Corsa, a well-known racing game renowned for its realistic vehicle dynamics and customization capabilities. Fig. \ref{fig:ac_system_diagram} shows our simulation environment architecture. A bi-directional interface based on the UDP communication and joystick emulator was implemented for the purpose of receiving live telemetry and transmitting user commands. The simulation vehicle models, including the engine, transmission and kinematics, were modeled based on AV-21, a full-scale autonomous race vehicle designed for the IAC. 

% To create realistic adversarial scenarios, we assumed that all vehicles on the track have the same engine power, wheel torque, and maximum speed. As a result, overtaking other vehicles was only possible if the opponent made mistakes or took advantage of the slipstream by closely following or staying ahead of other vehicles. This mirrors the professional human race, making the simulation more realistic.

\subsection{Dataset and Training}
We collected a total of 60 hours of simulation data, resulting in 120 GB of data samples. Humans and built-in AI drove the simulated AV-21 platform to gather demonstration trajectories across various race circuits, including Monza and the Las Vegas Motor Speedway. Of this data, 20\% was randomly selected as the validation set. Telemetry was updated at a rate of 100 Hz, and the input window sizes for the past trajectories of the ego vehicle and surrounding opponents were set to 0.5 seconds. To mitigate network overfitting, we adopted the motion dropout technique proposed in \cite{bansal2018chauffeurnet}. Each branch of the encoder module in the trajectory-level policy model (TPP) has a backbone of MobileNetV3-small \cite{howard2019searching}, except for the first and last layers, which were adjusted to match our input and output size of $1\times128$. We trained the hierarchical policies on a desktop with an i7-8700 CPU and a GTX 3090Ti GPU, using a batch size of 512.

\section{Evaluation Results}

\subsection{Baselines and Evaluation Metrics}
We re-implemented three relevant baselines, with a specific emphasis on the trajectory planning task. We selected baselines proposed that explicitly or implicitly output their confidence or uncertainty for overtaking trajectory planning in high-speed autonomous driving. Note that we excluded works that directly output control inputs \cite{fuchs2021super,wurman2022outracing,weiss2020deepracing}.
% We re-implemented three relevant baselines, with a specific emphasis on the trajectory planning task. We chose baselines that were proposed for overtaking trajectory planning for high-speed autonomous driving while explicitly/implicitly outputting their confidence or uncertainty. Note that we exclude some works which directly output control inputs \cite{fuchs2021super,wurman2022outracing,weiss2020deepracing}.

Baselines were trained and reproduced using the original works' hyperparameter setups. When the original configurations were not available, we used the same training settings as our approach. Below is a brief description of each baseline:
% Baselines were trained and reproduced following the original works' hyperparameter setups when possible. Otherwise, we used the same training configuration with ours. The following is a brief description of each baseline:
\begin{itemize}
  \item MTP-DCN \cite{djuric2020uncertainty}: A deep convolutional network with LSTM layers that predicts short-term vehicle trajectories while accounting for the uncertainty of motion.
  \item CSP-LSTM \cite{deo2018convolutional}: Uni-modal version of the network model that directly outputs the prediction uncertainty.
  \item DIRL \cite{jung2021incorporating}: Deep inverse reinforcement learning (DIRL) recovers the reward function behind demonstrations. The final trajectory with uncertainty is obtained through value iteration and state visitation frequency estimation.
\end{itemize}

We adopted two different evaluation metrics for trajectory planning and defined as follows:
\begin{itemize}
  \item Root Mean Squared Error (RMSE): represented the geometric distance between demonstrations and planned with the learned policy. %The lower the value, the closer the distance between the paths.
  \item Negative Log-Likelihood (NLL): represents the probability of the demonstration under the learned policy. The lower the value, the higher the probability.
\end{itemize}

\def\arraystretch{1.1}
\begin{table}[t]
\caption[Comparison results between multiple baseline policy models for trajectory planning.]{Comparison results between multiple baseline policy models for trajectory planning.}
\centering
\label{tab:comparison_evaluation}
\begin{adjustbox}{width=0.48\textwidth}
\begin{tabular}{ccccccc}
\hline
\multirow{2}{*}{\begin{tabular}[c]{@{}c@{}}Evaluation\\ Metric\end{tabular}} &
  \multirow{2}{*}{\begin{tabular}[c]{@{}c@{}}Planning\\ Horizon [sec]\end{tabular}} &
  \multirow{2}{*}{\begin{tabular}[c]{@{}c@{}}Race\\ Type\end{tabular}} &
  \multicolumn{4}{c}{Model} \\
  \cline{4-7} 
                      &  &  & BC & MTP-DCN & MC-DIRL & Ours  \\ \hline \hline
\multirow{4}{*}{\makecell{RMSE\\(m)}} & 1   & \multirow{2}{*}{\makecell{Solo\\Lap}}        
                        & 1.86 & 1.37    & 1.45   & \textbf{1.20} \\
                        & 1.5   &
                        & 3.28 & 3.12    & \textbf{2.99}   & 3.12 \\ \cline{2-7} 
                        & 1   & \multirow{2}{*}{\makecell{Four\\Players}}            & 5.01 & 3.98    & 3.39   & \textbf{3.39} \\
                        & 1.5   &                                     & 6.05 & 4.29    & 4.22   & \textbf{3.92} \\ \hline
\multirow{4}{*}{NLL}  & 1.0 & \multirow{2}{*}{\makecell{Solo\\Lap}}
                        & -  & 3.45     & 3.09    & \textbf{2.12}  \\
                        & 1.5 &                                       & -   & 4.12     & 3.46    & \textbf{2.92}  \\ \cline{2-7}  
                      & 1.0 & \multirow{2}{*}{\makecell{Four\\Players}}            & -  & 4.00     & 3.45    & \textbf{2.19}  \\
                      & 1.5 &                                       & -  & 4.37     & 3.65    & \textbf{2.71}  \\ \hline
\end{tabular}
\end{adjustbox}
\vspace{-0.5em}
\end{table}

% TODO, add following
% For the CSP-LSTM, we used the uni-modal version in the original paper. For DIRL, we visualized the state visitation frequency(SVF). For our proposed method, we ran five times optimization using Adam (Adaptive Moment Estimation) gradient descent optimization algorithm to find the maximum imitation prior trajectory starting with a random trajectory. The activation color is represented based on the variances for each trajectory point. Therefore, closer to the yellow color, the network predicts narrower variance.  

\subsection{Quantitative Evaluation}
We separately evaluated policies at different abstraction levels.
Table \ref{tab:comparison_evaluation} presents the performance comparison results of the TPP network across various scenarios. In solo lap scenarios (i.e., with no other agents on the track), all baselines, including our proposed method, showed good performance, likely due to the race line input guiding the policy effectively. We also evaluated performance in multi-player scenarios by manually selecting cases where four surrounding vehicles were within 50 meters of the ego vehicle. The results indicate that all baselines, including our proposed method, had increased errors in both evaluation metrics compared to the solo lap scenario. However, our model demonstrated the smallest errors across all planning horizons and the least performance degradation relative to other baselines. Additionally, our method exhibited significantly lower NLL error than the baselines, suggesting that the planned trajectory more closely resembled the expert's demonstration.
% Table \ref{tab:comparison_evaluation} presents the performance comparison results of the TPP network under various scenarios. In solo lap scenarios (i.e., with no other agents on the track), all baselines, including our proposed method, showed good performance. It is reasonable to assume that the race line input guided the policy to plan accordingly. We also evaluated the performance under multi-player scenarios. For this evaluation, we manually selected cases where four surrounding vehicles were present within 50 meters of the ego location. The results indicate that all baselines, including our proposed method, displayed increasing errors in both evaluation metrics compared to the solo lap scenario. Nevertheless, our model demonstrated the smallest error across all planning horizons and the least performance degradation compared to the other baselines. Moreover, our method exhibited a significantly lower NLL error compared to the baselines, implying that the planned trajectory was more similar to that of the expert's demonstration.

\def\arraystretch{1.1}
\begin{table}[t]
\caption[Evaluation results of RCP.]{Evaluation results of RCP.}
\centering
\label{tab:control_evaluation}
\begin{adjustbox}{width=0.48\textwidth}
\begin{tabular}{ccccc}
\hline
 &
  \multicolumn{2}{c}{\begin{tabular}[c]{@{}c@{}}Forward Controller \\w/o RCP Module\end{tabular}} &
  \multicolumn{2}{c}{\begin{tabular}[c]{@{}c@{}}Forward Controller \\w/ RCP Module\end{tabular}} \\ \hline 
\begin{tabular}[c]{@{}c@{}}Reference\\ Path\end{tabular} & Raceline & Centerline & Raceline & Centerline \\ \hline \hline
Avg tracking error [m] & 1.88 & 2.18 & \textbf{1.27} & 1.58 \\ \hline
Avg lap time [sec] & 38.19 & 38.72 & \textbf{37.91} & 38.32\\ \hline
\end{tabular}
\end{adjustbox}
\vspace{-1.5em}
\end{table}

% Notably, our proposed method outperformed the other baselines in terms of the performance index in most cases.

For the RCP model, we used two fixed trajectories: a pre-calculated race line and the track centerline. Table \ref{tab:control_evaluation} shows that RCP improves control performance in tracking accuracy and lap time, regardless of the path's geometry. With an already well-tuned forward controller and near-maximum vehicle speed, RCP provides a simple and effective way to refine the controller using imitation learning paradigm.

To further analyze our design, we conducted ablation studies. We maintained the TPP model architecture and evaluated the planning performance according to different input setups and planning horizons. Table \ref{tab:ablation-timeinterval} shows that providing the race line as an input improved the performance in terms of the RMSE performance index. However, longer or denser past trajectory inputs did not always improve the performance. We observed that longer or denser past trajectory inputs had a more significant impact on performance when the race line was not given as an input. These findings suggest that our planning model utilizes the past trajectory as more useful contextual cues to plan the trajectory when there is no explicit reference trajectory (i.e. race line).

\begin{figure*}[t]
    \centering
    \includegraphics[width=0.85\textwidth]{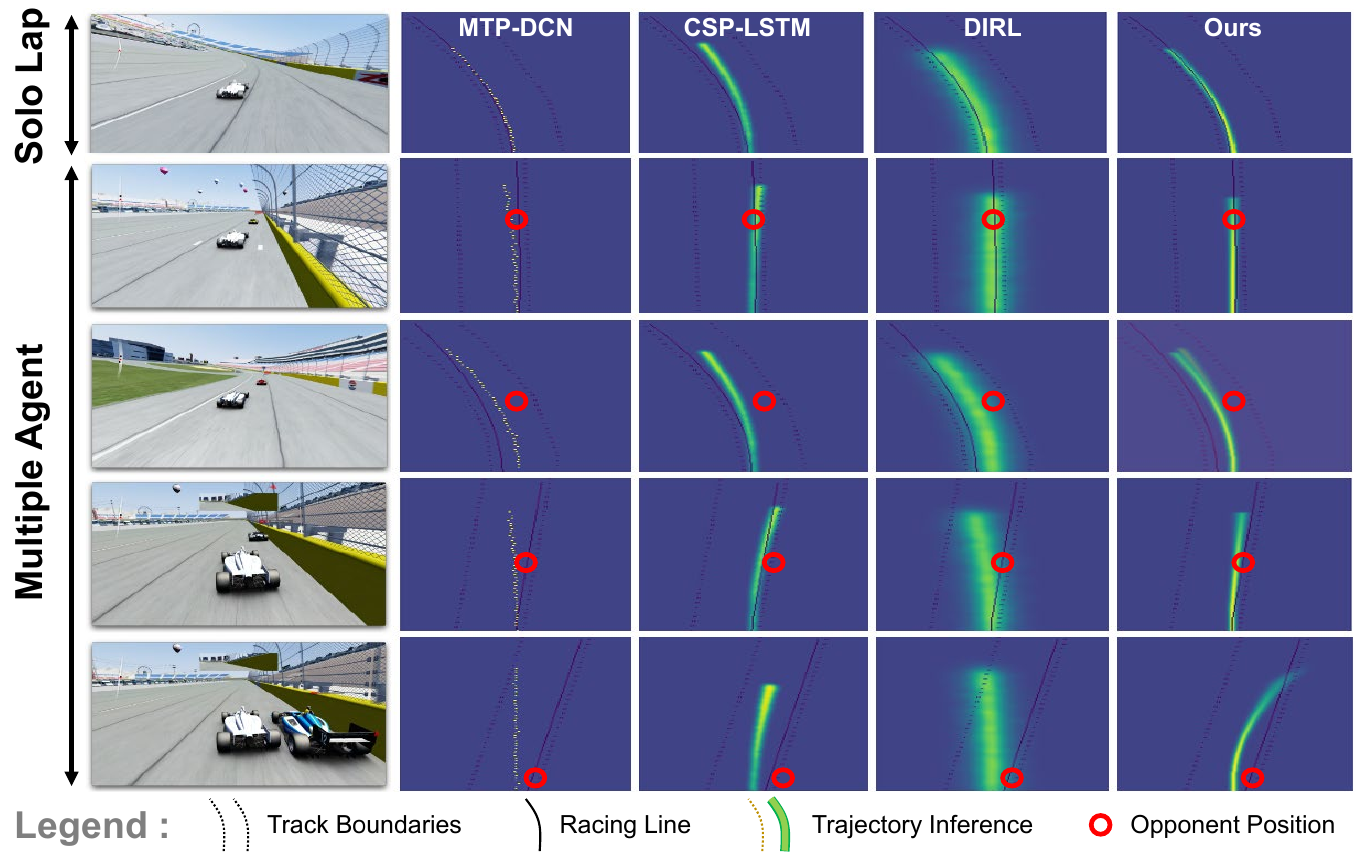}
    \caption{Qualitative evaluation of trajectory planning in solo (first row) and multi-agent (last four rows) racing scenarios using open-loop simulation. In solo racing, ideal outputs should have a high mean and low variance along the optimal racing line. In multi-agent scenarios, the trajectory policy should understand the surrounding circumstances to maximize progress and avoid collisions.}
    \label{fig:open-loop}
    \vspace{-0.5em}
\end{figure*}

\begin{figure*}[t!]
\centering
   \includegraphics[width=0.85\linewidth]{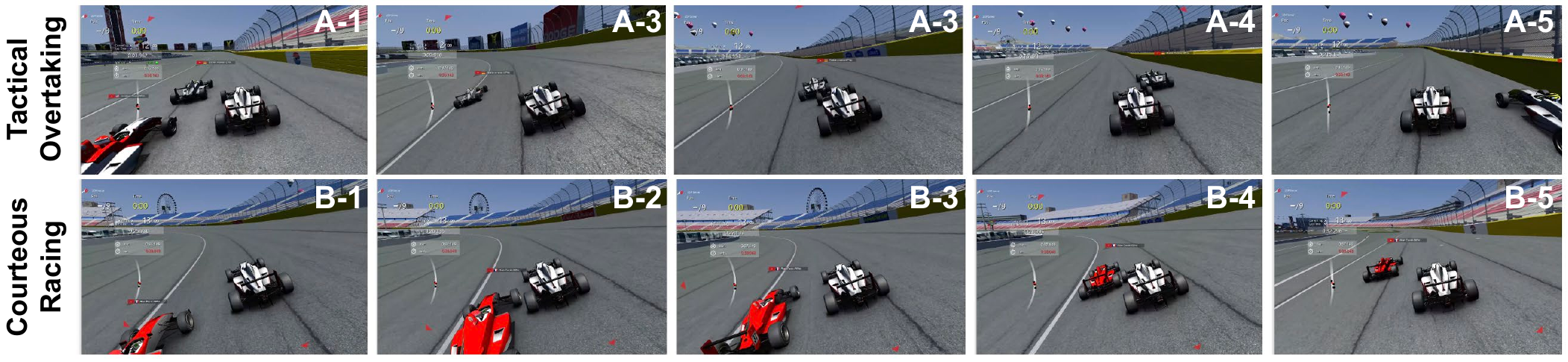}
   \label{fig:closed-loop}
    \caption[]{
    Closed-loop simulation results in tactical overtaking and courteous racing scenarios.
    % (Top row) The ego vehicle was driving on the race line towards the inner side of the track. A faster rear red opponent aggressively approached our rear, and our model yielded the race line to the opponent with the right amount of distance. (Bottom row) Two neighboring opponents were driving close to the ego vehicle, causing it to deviate from the race line. As a result, the gap between the ego vehicle and the ahead opponent increased. The ego vehicle then followed the ahead vehicle and overtook it to the left when it was faster and in close proximity.
    % Closed-loop simulation results in time series. (Top row) The ego vehicle was driving on the race line towards the inner side of the track. A faster rear red opponent aggressively approached our rear, and our model yielded the race line to the opponent with the right amount of distance. (Bottom row) Two neighboring opponents were driving close to the ego vehicle, causing it to deviate from the race line. As a result, the gap between the ego vehicle and the ahead opponent increased. The ego vehicle then followed the ahead vehicle and overtook it to the left when it was faster and in close proximity.
    }
    \label{fig:closed-loop}
    \vspace{-1.5em}
\end{figure*}

% Remove the following contents cuz of page limit
% We also studied the performance with four different dropout probabilities on the past trajectory for TPP, and the results are listed in Table \ref{tab:ablation-dropout}. Our findings revealed that utilizing past trajectory dropout improves performance to some extent, which is consistent with the findings in \cite{djuric2020uncertainty} from the learning-based planning paradigm perspective.

\def\arraystretch{1.1}
\begin{table}[t]
\caption[]{
Ablation study results using different inputs.
% Ablation study results on input past horizon, length, and race line.
}
\label{tab:ablation-timeinterval}
\centering
\begin{adjustbox}{width=0.45\textwidth}
\begin{tabular}{ccccccc}
\hline
\multirow{2}{*}{\begin{tabular}[c]{@{}c@{}}Evaluation\\ Metric\end{tabular}} &
  \multirow{2}{*}{\begin{tabular}[c]{@{}c@{}}Planning\\ horizon [sec]\end{tabular}} &
  \multirow{2}{*}{\begin{tabular}[c]{@{}c@{}}Raceline \\ Input\end{tabular}} &
  \multicolumn{3}{c}{Time Interval [sec]}\\
  \cline{4-6} &  &  & 0.01 & 0.02 & 0.05  \\ \hline \hline
\multirow{4}{*}{\makecell{RMSE\\(m)}}
& 0.5 & \multirow{2}{*}{\makecell{ \\ \checkmark}} & 4.39 & 4.47 & 4.12 \\
& 0.5 & & 3.96 & \textbf{3.59} & 3.60 \\ \cline{2-6}
& 1.0   & \multirow{2}{*}{\makecell{ \\ \checkmark}} & 4.00 & 4.12 & 3.92    \\
& 1.0   & & 4.02 & 3.67 & 3.64 \\ \hline
\end{tabular}
\end{adjustbox}
\vspace{-2.0em}
\end{table}

\subsection{Qualitative Evaluation}
\subsubsection{Open-loop simulation}

Fig. \ref{fig:open-loop} shows the open-loop simulation results with baselines. The first row illustrates the results during the solo lap scenarios where the policy is expected to predict a trajectory with high mean and low variance density along the race line, represented by the solid purple line. All baselines, including our method, predicted the future trajectory around the globally optimized race line. Except for MTP-DCN, other methods represented their outputs as predictive distributions, and all methods reasonably planned the trajectory in both straight and turn cases under solo lap scenarios. In particular, our method predicted the most accurate probability distribution close to the optimal race line until the end of the planning horizon.

The second-to-last row shows the prediction results under multi-player scenarios.
The outputs of MTP-DCN became jerky compared to the solo lap cases and predicted trajectories that failed to understand the racing context, going outside the track boundary while attempting to overtake opponents. Similarly, CSP-LSTM's output crossed the right track boundary and failed to consider environmental contexts.
Both DIRL and our method predicted staying behind the leading vehicle to maximize the slipstream rather than initiating an overtaking maneuver (second row). However, when the distance gap decreased and a speed advantage was gained, our method began predicting the possibility of overtaking on the right while still closely adhering to the racing line (third row).
% The second to the last row show the prediction results under multi-player scenarios. The outputs of MTP-DCN became jerky compared to the solo lap cases and predicted a trajectory to overtake the ahead vehicles to the left, which was out of the track boundary. The output of CSP-LSTM also crossed the right track boundary and failed to understand environmental contexts. DIRL and our method predicted staying behind the ahead vehicle to maximize the slipstream rather than initiate overtaking (Case 2). However, when the distance gap became smaller and had the speed advantage, our method started to predict the possibility of overtaking to the right while still trying not to deviate from the racing line (Case 3).

The fourth and fifth rows illustrate out-of-distribution examples during validation where the leading vehicle is completely stopped. MTP-DCN, CSP-LSTM, and DIRL failed to properly plan a collision avoidance trajectory. In contrast, our method predicted a trajectory that could quickly return to the racing line while considering the dynamics of a high-speed vehicle. However, at the end of the planning horizon in Case 5 (fifth row), our prediction slightly exceeded the right track boundary with very low density, which could be corrected in the subsequent closed-loop planning steps.
% The fourth and fifth rows illustrate Out-of-Distribution (OOD) examples during validation where the ahead vehicle is completely stopped. MTP-DCN, CSP-LSTM, and DIRL failed to plan the collision avoidance trajectory properly. On the other hand, our method predicted the trajectory that could quickly return to the race line while considering the dynamics of the high-speed vehicle. At the end of the planning horizon in Case 5, our prediction exceeded the right track boundary with a very low density.

\subsubsection{Closed-loop simulation} 
For challenging multi-agent scenarios, we spawned ten opponents on the track. As mentioned in Section \ref{sec:experiments}, all vehicles have identical specifications, and the game's built-in agents run at full throttle in most situations due to the oval track’s characteristics, making the racing scenarios more realistic and challenging. Note that our adversarial driving setup closely resembles human racing, where simple \textit{push to pass} maneuvers are not feasible. Demonstrations of fully autonomous racing with multiple agents are available as a video at \href{https://www.youtube.com/watch?v=7H6aTyUb7gE}{https://www.youtube.com/watch?v=7H6aTyUb7gE}.
% We spawned ten opponents on the track and inputted the closest three opponents' statuses for the network. As we introduced in Sec. \ref{sec:experiments}, all vehicles have the same vehicle spec, and the game's built-in agents run full-throttle in almost all situations according to the characteristics of the oval track. Note that our adversarial driving setup is very similar to the race of humans, where a simple \textit{push to pass} is impossible. Demonstration of the fully autonomous race with multiple players is uploaded as a video at \href{https://youtu.be/WBPczJbEWIE}{https://youtu.be/WBPczJbEWIE}.

Fig. \ref{fig:closed-loop} depicts the driving scenes during the closed-loop simulation in a time series. In the top row of Fig. \ref{fig:closed-loop}, two opponents were driving in close proximity to our ego vehicle. Since the raceline was occupied by the opponents, our vehicle could not drive on it and instead followed the center line of the track (A-1). As soon as there was enough space in front of the ahead vehicle, our vehicle moved behind it to take advantage of the slipstream (A-2 and A-3). After catching up with the opponent, our model planned the trajectory to overtake to the left (A-4 and A-5) when it had a sufficient speed advantage to safely pass.

In the bottom row of Fig. \ref{fig:closed-loop} scene B-1, the ego vehicle was driving on the race line towards the inner side of the track. The left rear red opponent was faster due to the slipstream effect and aggressively approached our rear (B-2). Although both vehicles almost touched, the opponent kept its position, and our ego vehicle had to give way to avoid a collision. In scene B-3, the ego vehicle moved to the center of the track to avoid unnecessary large deviations from the optimal race line. After the opponent passed, our vehicle followed closely behind it (B-4 and B-5) to catch up. These results demonstrate that our model actively interacts with neighboring agents' nuanced reactions while balancing performance and courtesy, similar to professional drivers.

\section{Conclusion}
\label{sec:discussion}
In this paper, we proposed the offline learning-based racing framework using hierarchical policy abstractions. Our trajectory planning policy integrates multiple contextual cues and predicts the likelihood of an expert prior via a density estimator. The trajectory output is then passed to a residual control policy designed to minimize the gap between a classical vehicle controller and expert demonstrations. Our approach was extensively evaluated against competing baselines in a realistic racing setup, demonstrating superior performance and the ability to overtake multiple agents while balancing agility and fairness.

Our study can be extended in two directions. The first is to train with demonstrations from professional-level drivers, enabling our framework to learn complex strategies and courteous behaviors, thereby enhancing its performance beyond human levels in multi-agent racing. The second direction is to implement a quantitative evaluation of courtesy in racing. Although defining and measuring courtesy is challenging, using real-world racing rulebooks could help assess the fairness of autonomous race cars with minimal reliance on human-engineered heuristics. We believe that our study, along with these extensions, could contribute not only to racing but also to urban highway scenarios, where vehicle interactions under high-speed conditions are crucial.
\bibliographystyle{IEEEtran} 
\bibliography{references}

\end{document}